\documentclass{article}

\usepackage{microtype}
\usepackage{graphicx}
\usepackage{subfigure}
\usepackage{booktabs} % for professional tables
\usepackage{amsmath, amsfonts, nccmath}
\usepackage{mathtools}
\usepackage{xcolor}
\usepackage{nicefrac}
\usepackage{enumitem}

\setlist{leftmargin=*,noitemsep}

\usepackage{hyperref}

\usepackage[accepted]{icml2020}

\DeclarePairedDelimiterX{\infdivx}[2]{(}{)}{%
  #1\;\delimsize\|\;#2%
}
\newcommand{\infdiv}{\mathcal{D}\infdivx}
\newcommand{\kldiv}{\mathrm{KL}\infdivx}

\icmltitlerunning{Model Fusion with Kullback--Leibler Divergence}

\begin{document}

\twocolumn[
\icmltitle{Model Fusion with Kullback--Leibler Divergence}



\icmlsetsymbol{equal}{*}

\begin{icmlauthorlist}
\icmlauthor{Sebastian Claici}{equal,mit,mitibm}
\icmlauthor{Mikhail Yurochkin}{equal,mitibm,ibm}
\icmlauthor{Soumya Ghosh}{mitibm,ibm}
\icmlauthor{Justin Solomon}{mit,mitibm}
\end{icmlauthorlist}

\icmlaffiliation{mit}{CSAIL, MIT, Cambridge, Massachusetts, USA}
\icmlaffiliation{ibm}{IBM Research, Cambridge, Massachusetts, USA}
\icmlaffiliation{mitibm}{MIT-IBM Watson AI Laboratory, Cambridge, Massachusetts, USA}

\icmlcorrespondingauthor{Sebastian Claici}{sclaici@csail.mit.edu}
\icmlcorrespondingauthor{Mikhail Yurochkin}{mikhail.yurochkin@ibm.com}

\icmlkeywords{Machine Learning, ICML}

\vskip 0.3in
]



\printAffiliationsAndNotice{\icmlEqualContribution} % otherwise use the standard text.
\newcommand{\Xmat}{\text{X}}
\newcommand{\Ymat}{\text{Y}}
\newcommand{\W}[1]{\mathcal{W}_{#1}}
\newcommand{\s}{\mathcal{S}}
\newcommand{\ind}[1]{\mathbf{1}[#1]}
\newcommand{\wvar}[1]{\tau_#1^{-1}}
\newcommand{\data}{\mathcal{D}}
\newcommand{\eye}{\mathbb{I}}
\newcommand{\T}{\mathcal{\theta}}
\newcommand{\Tau}{\mathcal{T}}
\newcommand{\blayer}{b_g}
\newcommand{\bnode}{b_0}
% Distributions
\newcommand{\invgamma}{\text{Inv-Gamma}}
\newcommand{\normal}{\mathcal{N}}
\newcommand{\halfcauchy}{C^{+}}
% Convenience
\newcommand{\taunode}{\tau_{kl}}
\newcommand{\tildetaunode}{{\tilde{\tau}}_{kl}}
\newcommand{\taulayer}{\upsilon_{l}}
\newcommand{\lambdanode}{\lambda_{kl}}
\newcommand{\lambdalayer}{\vartheta_{l}}
\newcommand{\lambdakappa}{\rho_{\kappa}}
\newcommand{\unode}{u_{kl}}
\newcommand{\wnode}{w_{kl}}
\newcommand{\betanode}{\beta_{kl}}
\newcommand{\w}{w_{ij,l}}
\newcommand{\half}{\frac{1}{2}}
\newcommand{\elbo}{\mathcal{L}}
\newcommand{\gradelbo}{\nabla_\phi\hat{\mathcal{L}}(\phi)}
\newcommand{\E}[1]{\mathbb{E}[#1]}
\newcommand{\Ewrt}[2]{\mathbb{E}_{#1}[#2]}
\newcommand{\ent}[1]{\mathbb{H}[#1]}
\newcommand{\real}[1]{\mathbb{R}^{#1}}
\newcommand{\bkappa}{b_\kappa}
\newcommand{\betamat}{\beta_{l}}
\newcommand{\mvn}{\mathcal{M}\mathcal{N}}
\newcommand{\nusample}{\nu_l^{(s)}}
\newcommand{\cbetamat}{M_{{\betamat}\mid\nu_l}}
\newcommand{\cU}{U_{{\betamat}\mid\nu_l}}
\newcommand{\vecB}{\vec{B}}
\newcommand{\vecM}{\vec{M}}
\newcommand{\cbetaj}{{\beta_j}\mid{\nu_j}}

\begin{abstract}
We propose a method to fuse posterior distributions learned from heterogeneous datasets. Our algorithm relies on a mean field assumption for both the fused model and the individual dataset posteriors and proceeds using a simple assign-and-average approach. The components of the dataset posteriors are assigned to the proposed global model components by solving a regularized variant of the assignment problem. The global components are then updated based on these assignments by their mean under a KL divergence. For exponential family variational distributions, our formulation leads to an efficient non-parametric algorithm for computing the fused model. Our algorithm is easy to describe and implement, efficient, and competitive with state-of-the-art on motion capture analysis, topic modeling, and federated learning of Bayesian neural networks.\footnote{Code link: \url{https://github.com/IBM/KL-fusion}}
\end{abstract}

\begin{figure}[!ht]
% \vskip 0.2in
\centering
\subfigure[Bayesian neural network trained on biased data]{\includegraphics[width=0.99\columnwidth]{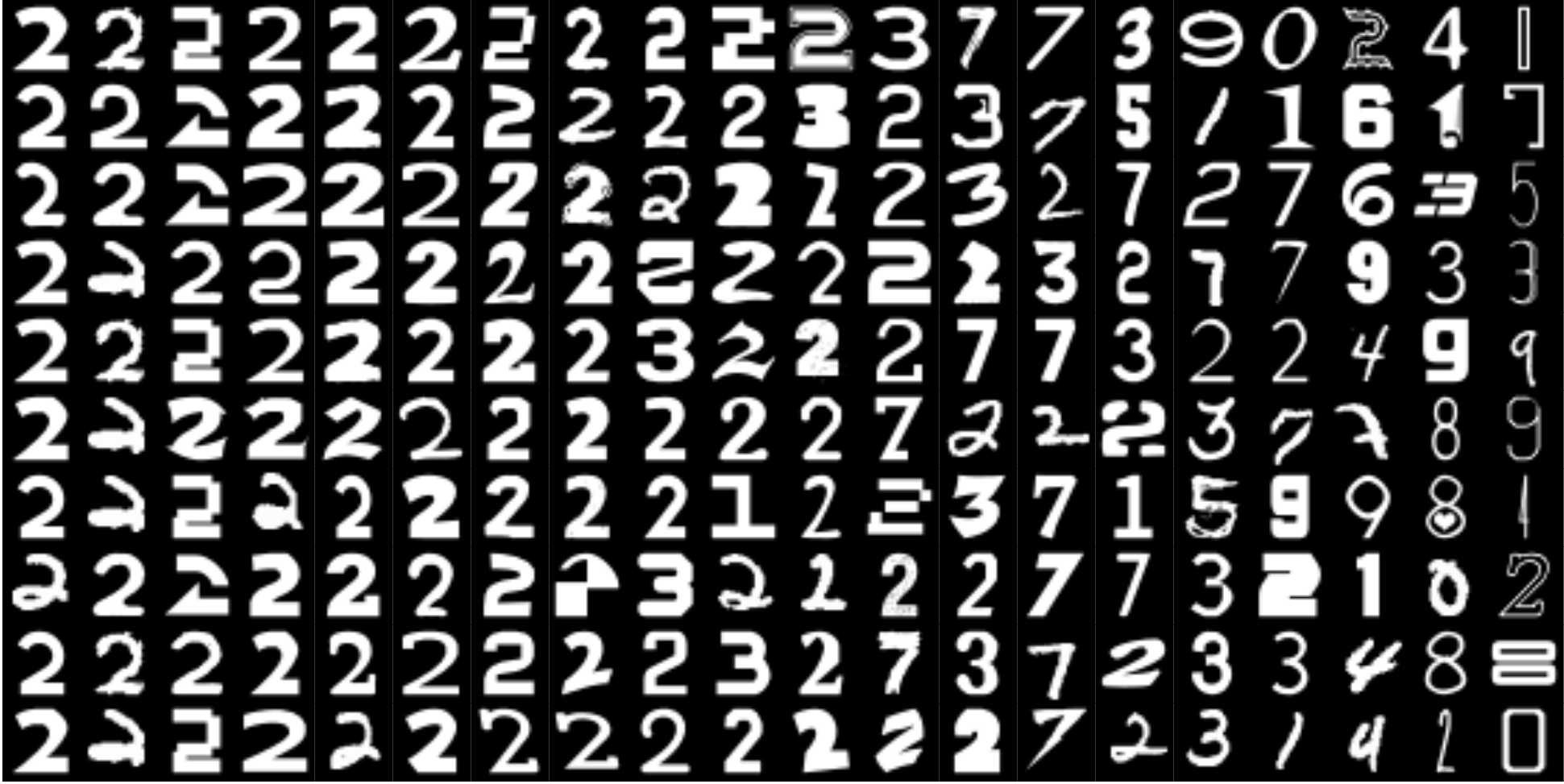}}
\subfigure[Fused Bayesian neural network]{\includegraphics[width=0.99\columnwidth]{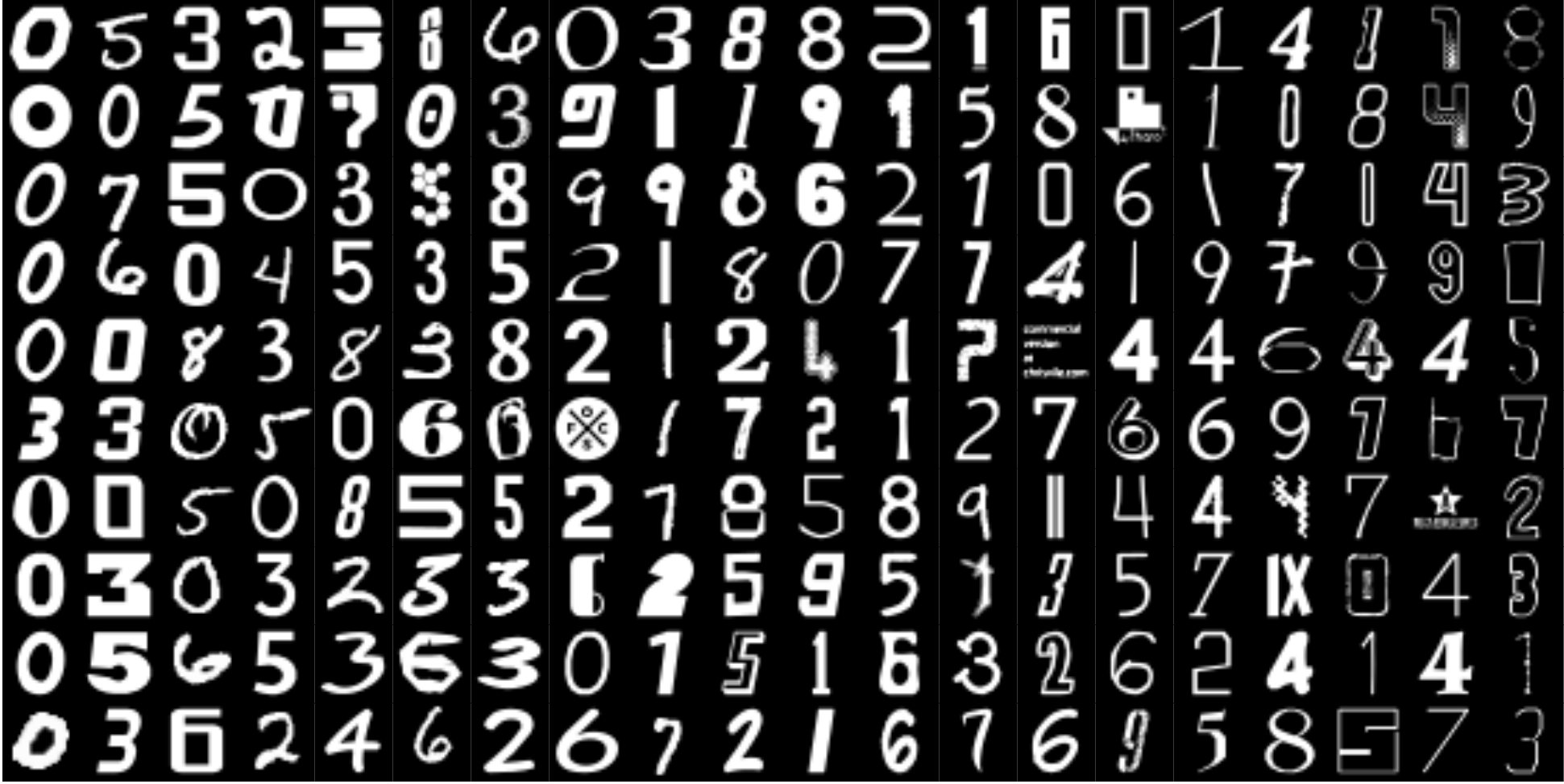}}
\vskip -0.1in
\caption{Bayesian neural networks can estimate how certain they are of each prediction they make. To illustrate this, we train a network on a subset of the MNIST dataset that mostly contains the digit $2$ and ask the network to predict on all digits; confidence is predictably low on digits other than $2$. However, if we are given several pre-trained networks that each exhibit certain biases, but which in aggregate have seen examples of all $10$ digits, can we merge them together into a network that has high accuracy and is confident on all of MNIST? We propose a \emph{model fusion} approach to solve this and other related problems. To illustrate our approach, we train Bayesian neural networks on subsets of MNIST that are skewed toward a few digits and fuse the trained networks. To show that the fused network is confident on all $10$ digits, we order samples by the entropy of the network's prediction for the fused model (b), and compare to a local model trained on mostly $2$s (a). These images sort examples about which the networks are most confident (left) to least confident (right). The fused network is still unconfident on out-of-sample data not used to train any of the fused networks---in this case, digits from fonts rather than handwriting---while the biased Bayesian network is unconfident on digits different than those seen in its training set.
\vspace{-.2in}
}
\label{fig:bnn_pics}
% \vskip -0.2in
\end{figure}

\section{Introduction}

In this paper, we study \emph{model fusion}, the problem of learning a unified global model from a collection of pre-trained local models. Model fusion provides a straightforward and efficient approach to \emph{federated learning} (FL), in which a model is learned from siloed data without direct access. 

As a motivating example, any one hospital may be able to use its patient data to train a model aiding diagnosis or treatment, but due to limited data and skew the resulting model may not be effective. To overcome this issue, a group of hospitals could in principle collaborate to produce a stronger model by pooling their data, but it is typically not permissible to share individual patient information between institutions. Federated learning---including model fusion---provides a means of generating a stronger or more widely-applicable model than what can be obtained from any one hospital's data, while only sharing aggregate information in the form of model parameters. 
As a second example, when learning from edge devices (e.g., next word prediction on smart phones), users often do not want their personal data to leave the device.  Federated learning algorithms only require parties (i.e., data owners) to share only local model parameters, rather than providing direct access to user-specific local data. % 
Some key aspects that distinguish federated learning from classical distributed learning are (1) constraints on the frequency of communication and (2) heterogeneity of the local datasets. 

To overcome these challenges, we fuse heterogeneous models in a \emph{single} communication round that aggregates locally-trained models into a global model. This ``one-shot'' approach, which distinguishes model fusion from other FL methods, is crucial for certain applications yet is largely overlooked with the exception of \citet{yurochkin2019spahm,yurochkin2019pfnm}. In particular, the one shot approach allows parties to erase data in favor of storing only their local models.

Many common examples motivating FL require well-calibrated uncertainty measurements for the fused model; our examples of medical decision-making and word prediction on smart phones provide applications in which uncertainty quantification can be used to avoid making an inaccurate and potentially unsafe intervention.  
This context motivates a \emph{Bayesian} approach to model fusion and FL more broadly. The Bayesian nature of our fusion algorithm bolsters trust in systems that use this machinery in applications that encounter out-of-distribution samples when deployed in practice. We demonstrate the value of a Bayesian approach to model fusion through applications from topic models to neural networks.

\paragraph{Contributions.} We present a Bayesian approach to model fusion, in which \emph{local} models trained on individual datasets are combined to learn a single \emph{global} model.  Our approach fuses models represented using the mean field approximation, common for lightweight and robust local estimation. It is nonparametric in the sense that it does not require the dimensionality of the fused mean field model to be fixed \emph{a priori}.  Specific contributions include:
\begin{itemize}
    \item a non-parametric method to determine the posterior distribution of the fused global model;
    \item an easily-implemented assign-then-average fusion procedure that scales to large numbers of local posterior distributions and fused model components;
    \item a model for fusion that is flexible enough to handle posterior distributions in any exponential family; and
    \item comprehensive validation demonstrating effectiveness for applications including motion capture analysis, topic modeling, and Bayesian neural networks.
\end{itemize}
\section{Related Work}

This paper develops  model fusion techniques for approximate Bayesian inference, combining parametric approximations to an intractable posterior distribution. While we primarily focus on mean-field variational inference (VI), owing to its popularity, our methods are equally applicable to Laplace approximations~\cite{bishop2006pattern}, assumed density filtering~\cite{opper1998bayesian}, and expectation propagation~\cite{minka2001expectation}  methods that learn a parametric approximation to the posterior. Variational inference seeks to approximate the true posterior distribution by a tractable approximate distribution by minimizing the KL divergence between the variational approximation and the true posterior. In contrast with Markov chain Monte Carlo methods, VI relies on optimization and is thus able to exploit advances in stochastic gradient methods allowing for VI based inference algorithms to scale to large data and models with a large number of parameters, such as  Bayesian Neural Networks (BNNs)~\cite{neal1995bayesian} considered in this work.

Distributed posterior inference has been actively studied in the literature \citep{hasenclever2017distributed, broderick2013streaming, bui2018partitioned, srivastava2015wasp, bardenet2017markov}. As with distributed optimization, however, the goal is typically to achieve computational speedups, leading to approaches ill suited for model fusion due to high number of communication rounds required for convergence and assumption on the homogeneity of the datasets. Moreover, the inherent permutation invariance structure of many high-utility models (e.g., topic models, mixture models, HMMs, and BNNs) is ignored by prior distributed Bayesian learning methods as it is of minor importance when many communication rounds are permissible. On the contrary, our model fusion formulation requires careful consideration of the permutation structure as we show in the subsequent section. Aggregation of Bayesian posteriors respecting permutation structure was considered in \citet{campbell2014approximate}, but their method is limited to homogeneous data and requires combinatorial optimization except few special cases. Subsequent work relaxes the homogeneity constraint and propose a greedy streaming approach for Dirichlet process mixture models \cite{campbell2015streaming}.

\citet{yurochkin2019spahm,yurochkin2019pfnm} studied fusion of parameters of permutation invariant models learned from heterogeneous data, however their approach is not suitable when a global posterior, rather than a point estimate, is desired. This limitation makes their methods ill-suited for fusion of Bayesian neural networks where posterior is required to assess prediction uncertainty---the key utility of BNNs. Their method also precludes using full information provided by the local posteriors, e.g.\ covariances, that may be necessary to efficiently identify global model with fusion as we demonstrate in the experimental studies.
\section{Homogeneous Fusion}
\label{sec:homogeneous-fusion}

Before introducing our non-parametric model for heterogeneous fusion, we consider the simpler \emph{homogeneous} fusion problem. The purpose of this section is to define notation and to introduce the building blocks for the algorithm of \S\ref{sec:heterogeneous-fusion}.  

Assume we have $D$ datasets on which we run some inference procedure to recover a mean-field approximation to the posterior distribution. For dataset $j$, we are thus given
\begin{align*}
    p_j(z_1, \ldots, z_{L}) = \prod_{l=1}^{L} q(z_l | \theta^j_{l})
\end{align*}
where $q(z_l | \theta^j_{l})$ is the approximate posterior of component $z_l$ parameterized by $\theta^j_{l}$. An example to keep in mind is topic modeling, where the $z_l$'s represent topics and the $\theta_l$'s are the posterior variational Dirichlet parameters for topic $l$. Our goal is to recover a single global distribution over the $z_l$'s without returning to the data to learn this global distribution. Hence, all we can use for inference are the parameters $\theta^j_l$ extracted from each dataset $j$. 

We can pose the problem of recovering a global posterior as that of minimizing a divergence $\infdiv{\cdot}{\cdot}$ to the local posteriors, but the ordering of posterior parameters is different from dataset to dataset. This phenomenon is called \textit{label switching}, and is caused by the permutation invariance of the posterior  \cite{monteiller2019alleviating}.

Because we are in the \emph{homogeneous} case, we can assume that the components of each local posterior can be put into correspondence across datasets. This allows us to assume that the global model admits the same product factorization:
\begin{align*}
    \bar{p}(z_1, \ldots, z_{L}) = \prod_{g=1}^{L} q(z_g | \bar{\theta}_g).
\end{align*}
Our goal is to find an effective choice of the $\bar{\theta}_g$'s, but we must be careful in how we define the objective.
In particular, the ordering of components for each dataset can be arbitrary as the posteriors are invariant to permutations in the parameters, and our objective function must account for this permutation invariance. 

To this end, we introduce auxiliary optimization variables $P^j$ for each dataset posterior. The $P^j$'s are permutation matrices that allow for parameter reorderings across local models. The problem we wish to solve is then:
\begin{equation}
\label{eq:fusion-homogeneous-objective}
    \begin{split}
        \min_{\{\bar{\theta}_g\}, \{P^j\}}\quad &\sum_{j=1}^D \infdiv*{ \prod_{l=1}^L q\left(z_g \left| \sum_{g=1}^L P^j_{lg} \bar{\theta}_{g}\right.\right)\!}{\prod_{l=1}^L q(z_l | \theta^j_l)}\\
        \text{subject to}\quad & \sum_{l=1}^L P^j_{lg} = 1,\, \sum_{g=1}^L P^j_{lg} = 1,\, P^j_{lg} \in \{0, 1\}.
    \end{split}
\end{equation}
The notation for indices here and later follows the convention that $g$ indexes global parameters, $l$ indexes local parameters, and $j$ indexes datasets; $\mathcal{D}$ is a divergence.
To read this equation, notice that the inner sum $\sum_{g=1}^L P^j_{lg} \bar{\theta}_{g}$ selects the global parameter that best explains the $z_l$. Thus, we can think of \eqref{eq:fusion-homogeneous-objective} as asking for the choice of global parameters $\bar{\theta}_g$ as well as a permutation for each dataset telling how to put the $\bar{\theta}_g$ and $\theta^j_l$ in correspondence such that the total divergence over all datasets is minimized.

The tractability of this problem depends on the divergence $\infdiv{\cdot}{\cdot}$. One choice that greatly simplifies the problem is the Kullback--Leibler ($\mathrm{KL}$) divergence, which decomposes over product distributions and allows us to write \eqref{eq:fusion-homogeneous-objective} as 
\begin{equation}
    \begin{split}
        \min_{\{\bar{\theta}_g\}, \{P^j\}}\quad &\sum_{j=1}^D \sum_{l=1}^L \kldiv*{q\left(z_g \left| \sum_{g=1}^L P^j_{lg} \bar{\theta}_{g}\right.\right)}{q(z_l | \theta^j_l)}\\
        \text{subject to}\quad & \sum_{l=1}^L P^j_{lg} = 1,\, \sum_{g=1}^L P^j_{lg} = 1,\, P^j_{lg} \in \{0, 1\}.
    \end{split}
\end{equation}
To further simplify, we can exploit the binary constraints in our problems. The $P^j$'s are binary matrices with a single $1$ in each row and column. Because all other entries of $P^j_{\cdot g}$ are $0$, we can move the sum outside the $\mathrm{KL}$ term, as $P^j_{lg} \cdot \kldiv{\cdot}{\cdot}$ will not contribute to the objective function if $P^j_{lg}=0$. The final form of our objective becomes
\begin{equation}
\label{eq:fusion-homogeneous-final}
    \begin{split}
        \min_{\{\bar{\theta}_g\}, \{P^j\}}\quad &\sum_{j=1}^D \sum_{l,g=1}^L P^j_{lg} \kldiv*{q(z_g | \bar{\theta}_g)}{q(z_l | \theta^j_l)}\\
        \text{subject to}\quad & \sum_{l=1}^L P^j_{lg} = 1,\, \sum_{g=1}^L P^j_{lg} = 1,\, P^j_{lg} \in \{0, 1\}.
    \end{split}
\end{equation}

Problem \eqref{eq:fusion-homogeneous-final} is easier to solve than what we started with in \eqref{eq:fusion-homogeneous-objective}. With fixed $\{P^j\}$, the problem is a barycenter or clustering problem under the $\mathrm{KL}$ divergence, which is known in closed form for exponential family distributions (\S\ref{sec:averaging}), while with fixed $\{\bar{\theta}_g\}$ it reduces to a stable marriage assignment that can be solved efficiently using the Hungarian algorithm \cite{kuhn1955hungarian}; this step has worst-case $\tilde{O}(L^3)$ complexity. Since the local parameters $\theta_l^j$ are independent across datasets, the $P^j$'s can be computed independently. 

\subsection{Averaging parametric distributions}\label{sec:averaging}

For posterior distributions that are in the same exponential family, computing their barycenter under the KL divergence amounts to averaging their natural parameters \cite{banerjee2005clustering}. 
In particular, given distributions $q_i$ in the same exponential family $Q$ with natural parameters $\eta_i$ as well as a set of weights $\lambda_i \geq 0$ with $\sum_i \lambda_i = 1$, the solution to
\begin{equation*}
    \min_{q\in Q} \sum_{i=1}^n \lambda_i \kldiv{q}{q_i}
\end{equation*}
is a distribution $q^*\in Q$ with natural parameter 
$\eta^* = \sum_{i=1}^n \lambda_i \eta_i$. In our case, given a assignments $\{P^j\}$, we can solve for $\bar{\theta}_g$ by minimizing
\begin{equation}
\label{eq:bregman-bary}
    \min_{q_g\in Q}\sum_{j=1}^D\sum_{l=1}^L P^j_{lg} \kldiv*{q_g}{q(z_l | \theta^j_l)}
\end{equation}
and converting from natural parameters to $\bar{\theta}_g$.
\section{Heterogeneous Fusion}
\label{sec:heterogeneous-fusion}

The homogeneous approach requires two assumptions that often do not hold: (1) that the local posterior distributions contain the same number of components and (2) that these components can be matched to one another. 

As an extreme example, if we run an inference procedure on data gathered from multiple hospitals, each specializing in a particular set of diseases, we ideally expect the combined model to incorporate information about \emph{all} diseases treated across \emph{all} hospitals. If we run the procedure on the data from each hospital individually, however, then each hospital's model only contains information about the diseases it treats; the mismatch between maladies at different hospitals prevents us from matching their parameters bijectively.

Motivated by this example, we observe that in practical settings the global fused model likely needs \emph{more} components than the local models to capture the fact that local data can be skewed or missing. We call this setting the \emph{heterogeneous} case. In the heterogeneous model, the inference procedure on each dataset may find a different number of components. Some components present in one dataset may not be present in another, and the total number of components is unknown---demanding a nonparametric solution.

\subsection{Heterogeneous Model}

To describe our model for heterogeneous fusion, we make a few notational changes. Let $G$ be an estimate of the number of global components (we will see how $G$ can be inferred in \S\ref{sec:estimate-global}), and $L_j$ be the number of components in dataset $j$. Instead of permutation matrices, the $P^j$ are singly-stochastic, since there may be unmatched global components. We can modify \eqref{eq:fusion-homogeneous-objective} to take these changes into account:
\begin{equation}
\label{eq:fusion-heterogeneous-objective}
    \begin{split}
        \min_{\{\bar{\theta}_g\}, \{P^j\}}\quad &\sum_{j=1}^D \infdiv*{ \prod_{l=1}^{L_j} q\left(z_g \left| \sum_{g=1}^{G} P^j_{lg} \bar{\theta}_{g}\right.\right)\!}{\prod_{l=1}^{L_j} q(z_l | \theta^j_l)}\\
        \text{subject to}\quad & \sum_{l=1}^{L_j} P^j_{lg} \leq 1,\, \sum_{g=1}^G P^j_{lg} = 1,\, P^j_{lg} \in \{0, 1\}.
    \end{split}
\end{equation}

If we think of $P^j$ as an $L_j\times G$ matrix, the inequality constraint in \eqref{eq:fusion-heterogeneous-objective} forces columns of $P^j$ to zero if global component $g$ is not used. The same simplifications we used to derive \eqref{eq:fusion-homogeneous-final} in \S\ref{sec:homogeneous-fusion} apply here, leading to the following formulation of the heterogeneous problem when $G$ is known:
\begin{equation}
\label{eq:fusion-heterogeneous-final}
    \begin{split}
        \min_{\{\bar{\theta}_g\}, \{P^j\}}\quad &\sum_{j=1}^D \sum_{l=1}^{L_j}\sum_{g=1}^G P^j_{lg} \kldiv*{q(z_g | \bar{\theta}_g)}{q(z_l | \theta^j_l)}\\
        \text{subject to}\quad & \sum_{l=1}^{L_j} P^j_{lg} \leq 1,\, \sum_{g=1}^G P^j_{lg} = 1,\, P^j_{lg} \in \{0, 1\}.
    \end{split}
\end{equation}
This model can cope with mismatched components among the local models (e.g., different diseases appearing at different hospitals), but it does not tell how to choose the number of global components $G$---a challenge we address next.

\subsection{Estimating the number of global components}
\label{sec:estimate-global}

A key issue with \eqref{eq:fusion-heterogeneous-final} is that we do not know the true number of global components $G$. If we na\"ively overestimate this parameter, there is neither a term in the objective nor a constraint in \eqref{eq:fusion-heterogeneous-final} that would reduce the number of components that are used; in the extreme case, the fused model would simply concatenate all the components of the local models without clustering any of them together. 

To motivate our approach to choosing $G$, we first consider the simpler problem of matching a \emph{single} local model to the global model. The optimization variables in this case are the $\bar{\theta}_g$'s as before, as well as a single $L\times G$ matching matrix $P$. We believe that the local model can be approximated best by a small number of components, and we wish to encode this explicitly in the objective. 

Recall that a component of the global model that goes unused corresponds to a $0$ column of $P$. For a binary matrix $P$ with $G > L$, $0$ columns occur for exactly $G - L$ of the columns when optimizing \eqref{eq:fusion-heterogeneous-final} with $L=1$.  But, if we relax the binary constraint, this may no longer be the case. Inspired by the approach of \citet{carli2013convex} to clustering using optimal transport, to promote $0$ columns in $P$ for the relaxed problem where $P_{lg} \in [0, 1]$, we regularize our problem using the $L_{2,1}$ matrix norm. This approach can be understood as optimizing the $L_1$ norm of the vector of $L_2$ column norms of $P$, promoting sparsity in the vector of norms and hence existence of $0$ columns in $P$. In mathematical notation, our relaxed objective with the sparsity-promoting regularizer is 
\begin{equation}
\label{eq:fusion-heterogeneous-one}
    \begin{split}
       \sum_{l,g} P_{lg} \kldiv*{q(z_g | \bar{\theta}_g)}{q(z_l | \theta_l)} +
        \lambda \sum_{g=1}^G \left(\sum_{l=1}^L P_{lg}^2 \right)^{\nicefrac{1}{2}}.
    \end{split}
\end{equation}
The mixed-norm regularizer is needed instead of a simpler $L_1$ regularizer, since the constraints of \eqref{eq:fusion-heterogeneous-final} effectively prescribe the $L_1$ norm of $P$ to a fixed constant.

A na\"ive extension to multiple local models might sum \eqref{eq:fusion-heterogeneous-one} over all datasets, but this approach only promotes sparsity within the individual global-to-local assignments. This can lead to a scenario wherein every global component is assigned to some local component at least once, again saturating the total number $G$ of available global components. Instead, following the intuition above, we can view the set $\{P^j\}_{j=1}^D$ as a tensor and minimize an $L_{2, 2, 1}$ tensor norm:
\begin{equation}
\label{eq:fusion-heterogeneous-regularization}
    \sum_{g=1}^G\left(\sum_{j=1}^D\left(\sum_{l=1}^{L_j} (P_{lg}^j)^2\right)\right)^{\nicefrac{1}{2}}.
\end{equation}
This quantity is the $L_{2, 1}$ norm of the $G\times D$ matrix whose element at position $(g, j)$ is the norm of column $g$ in $P^j$. 

\subsection{The heterogeneous matching problem}\label{sec:algorithm}
Combining \eqref{eq:fusion-heterogeneous-objective}--\eqref{eq:fusion-heterogeneous-regularization} yields the following problem:
\begin{equation}
\label{eq:fusion-heterogeneous-complete}
    \begin{split}
    \min_{\{\bar{\theta}_g\}, \{P^j\}}\: &
    \begin{aligned}[t]
        \sum_{j=1}^D \sum_{l=1}^{L_j}\sum_{g=1}^G P^j_{lg} &\kldiv*{q(z_g | \bar{\theta}_g)}{q(z_l | \theta^j_l)} +\\ 
        & \lambda \sum_{g=1}^G\left(\sum_{j=1}^D\left(\sum_{l=1}^{L_j} (P_{lg}^j)^2\right)\right)^{\nicefrac{1}{2}}
    \end{aligned}\\   
    \text{subject to}\: & \sum_{l=1}^{L_j} P^j_{lg} \leq 1,\, \sum_{g=1}^G P^j_{lg} = 1,\, P^j_{lg} \in \{0, 1\}.
    \end{split}
\end{equation}

\paragraph{Choosing $\lambda$.} To choose a regularization parameter $\lambda$, we ensure that the two terms in the objective have the same scale. Since the $P^j$'s are positive matrices with entries less than $1$, the scale of the problem is given by the $\kldiv{\cdot}{\cdot}$ terms. An empirically effective choice is to scale the divergences by their standard deviation and set $\lambda = 0.1$.

\paragraph{Alternation algorithm.}
Optimizing \eqref{eq:fusion-heterogeneous-complete} is not straightforward. In the homogeneous case, we exploited the fact that $P^j$ was a permutation to relax the binary constraints and recover a binary matrix; thanks to the inequality in \eqref{eq:fusion-heterogeneous-final}, however, we are no longer guaranteed to find a binary $P^j$ if we relax the $P^j_{lg}\in\{0,1\}$ constraint. Relaxing the binary constraints, however, turns \eqref{eq:fusion-heterogeneous-complete} into a convex problem in the $\{P^j\}$ and $\{\bar{\theta}_g\}$ individually. Our alternation approach from \S\ref{sec:homogeneous-fusion} can be modified to suit our new problem. Because we are no longer guaranteed that the $P^j$'s are permutations, we take a weighted average when updating the $\bar{\theta}_g$'s using \eqref{eq:bregman-bary}.
\begin{figure*}[ht]
% \vskip -0.2in
\begin{center}
\subfigure[Hausdorff distance estimation error]{\includegraphics[width=0.49\textwidth]{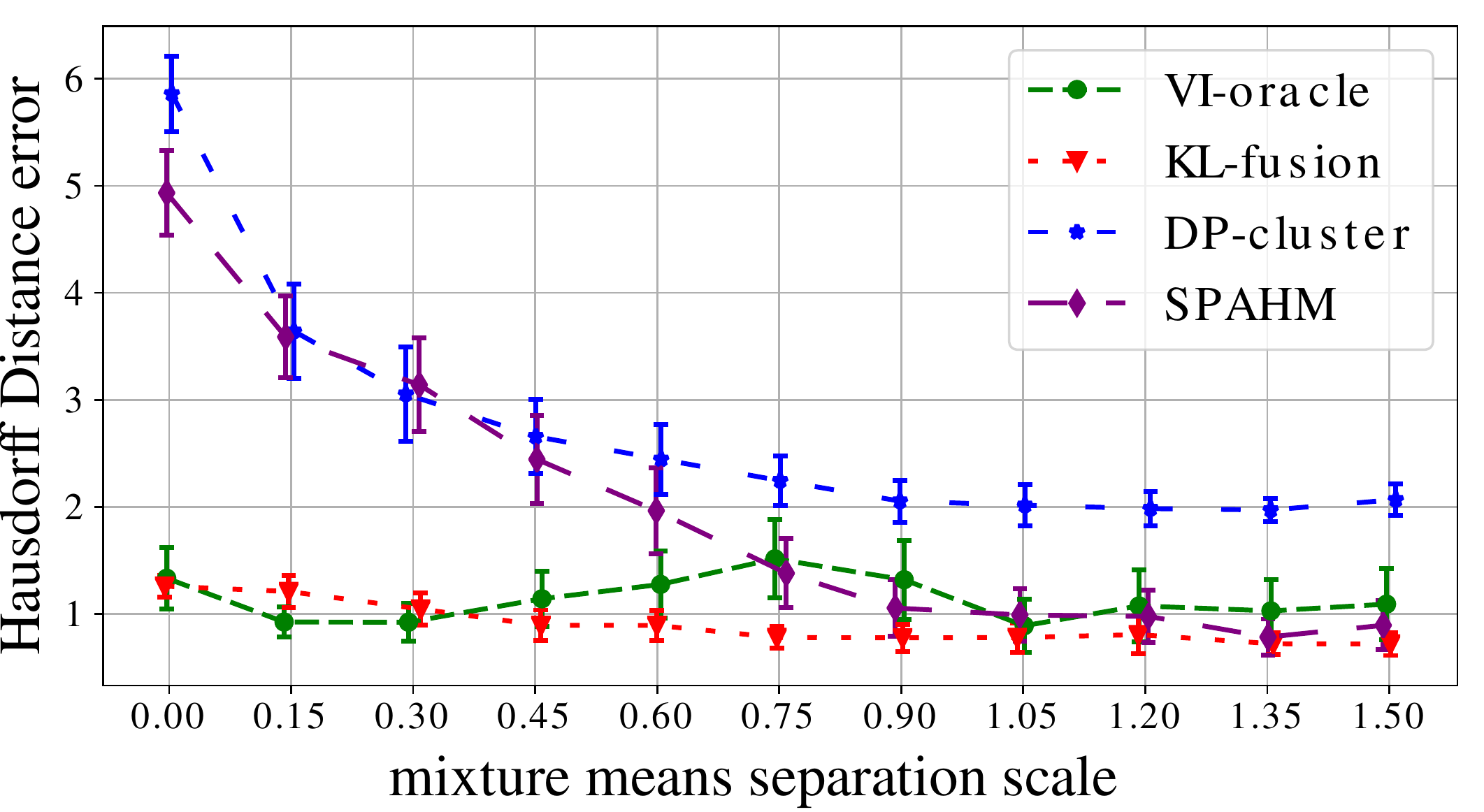}}
\subfigure[Global model size estimation error]{\includegraphics[width=0.49\textwidth]{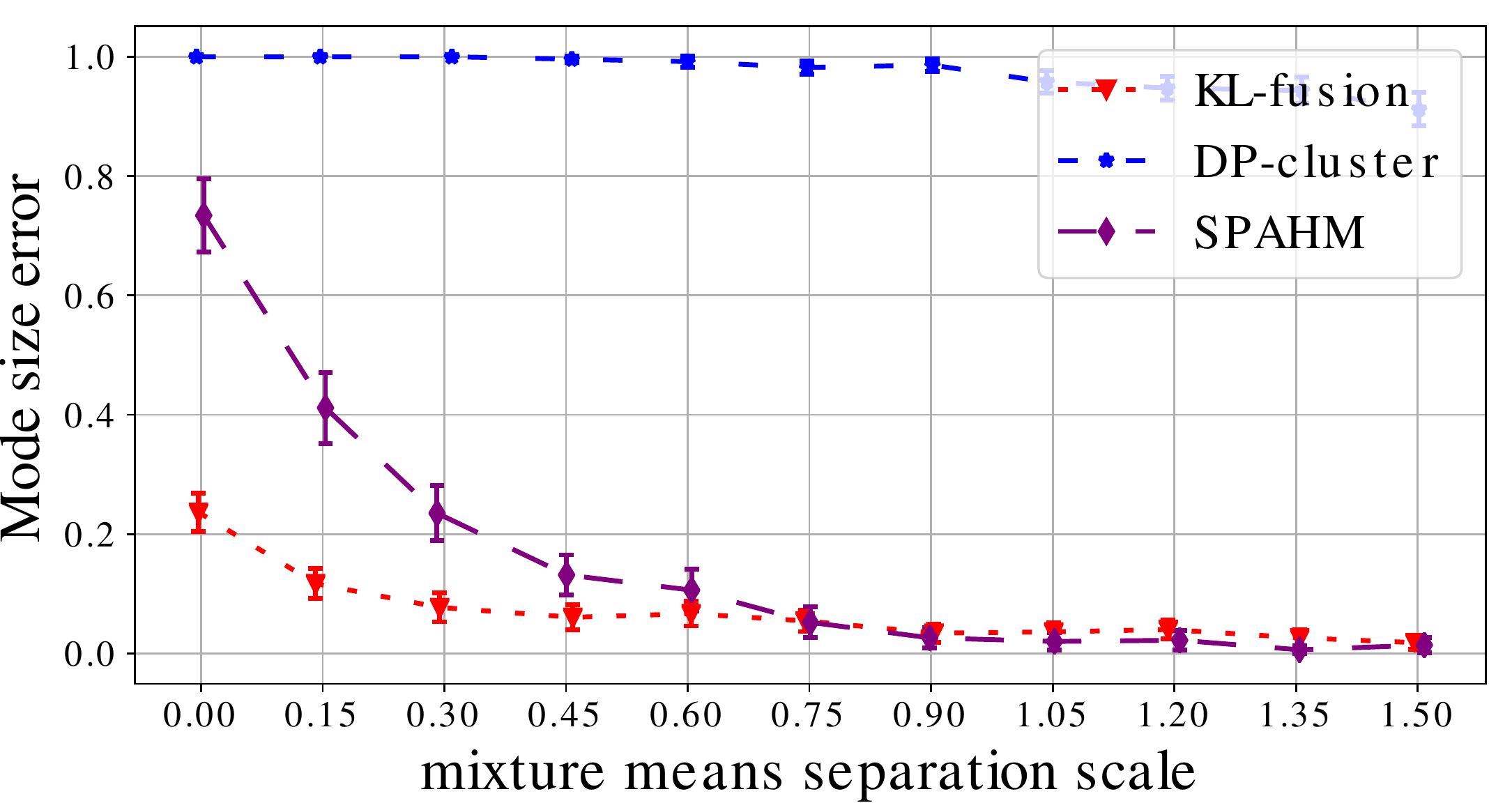}}
\vskip -0.1in
\caption{Fusion of Gaussian mixture posteriors under varying degree of separation between data generation mixture components. KL-fusion can identify true mean parameters under the low separation regime utilizing the covariance information.\vspace{-.15in}}
\label{fig:separation}
\end{center}
% \vskip -0.2in 
\end{figure*}

\begin{figure*}[h]
% \vskip 0.2in
\begin{center}
\subfigure[Hausdorff distance estimation error]{\includegraphics[width=0.49\textwidth]{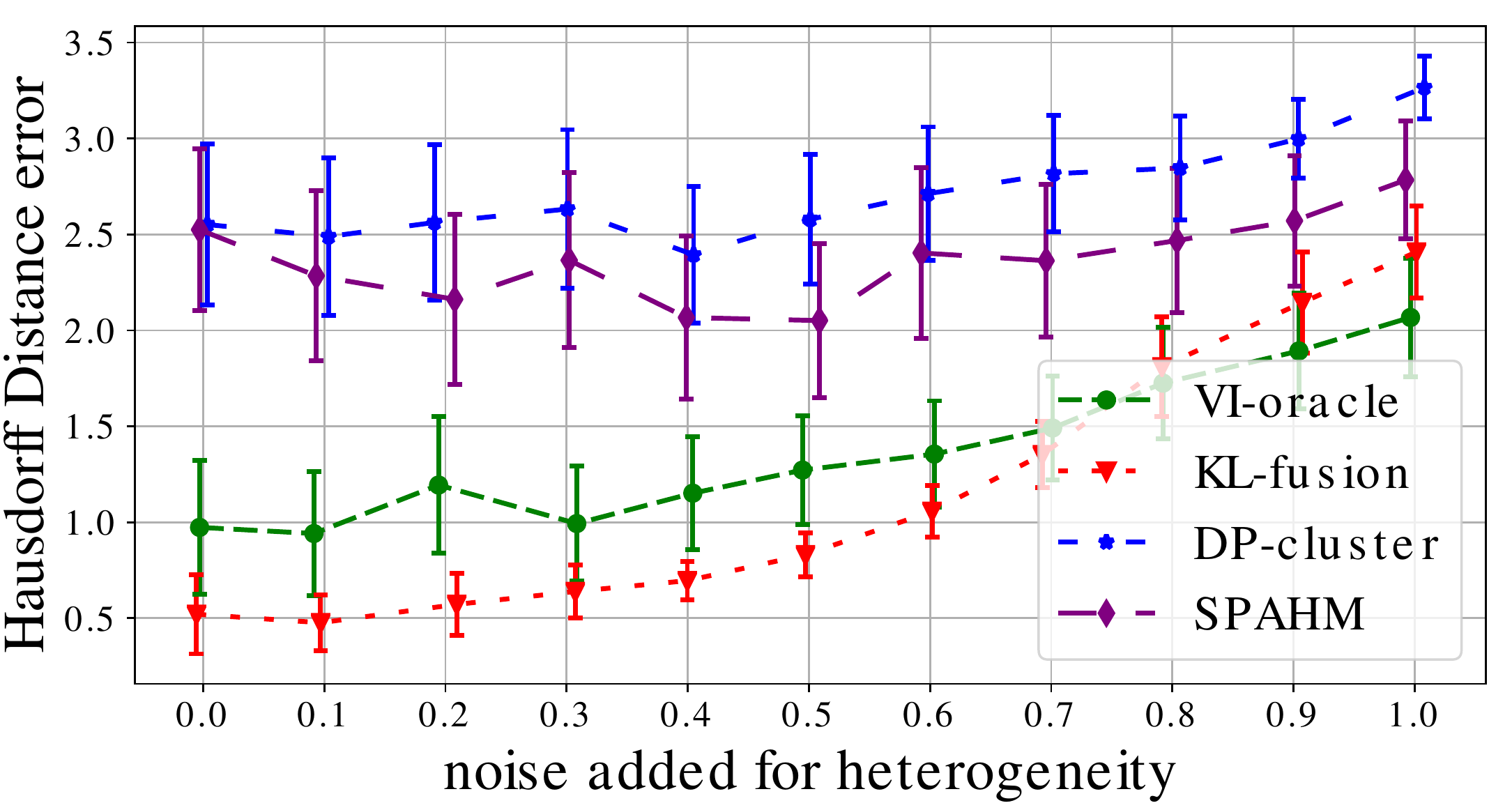}}
\subfigure[Global model size estimation error]{\includegraphics[width=0.49\textwidth]{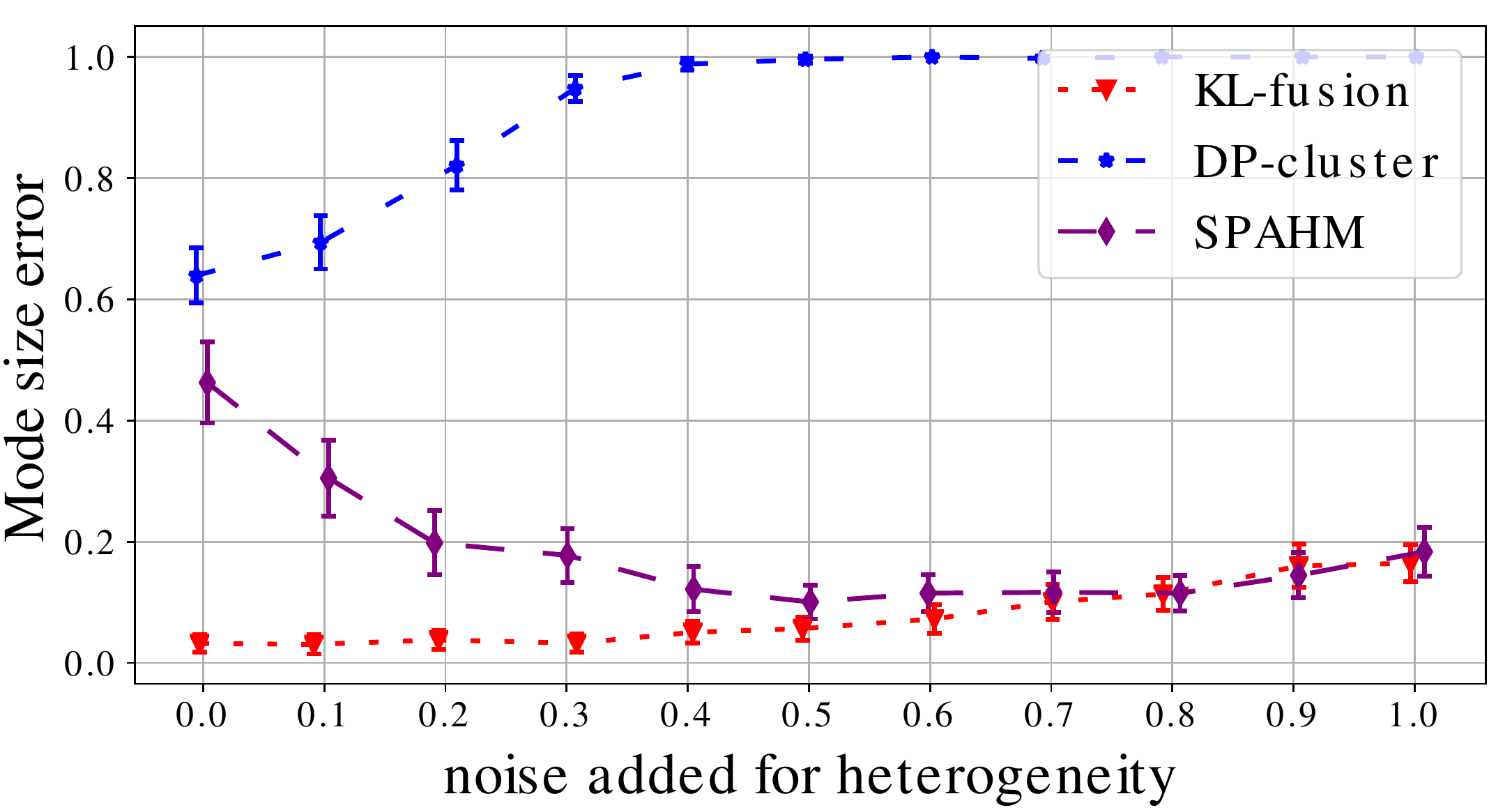}}
\vskip -0.1in
\caption{Fusion of Gaussian mixture posteriors under varying heterogeneity in local datasets enforced via noise standard deviation added to global means when simulating local means. KL-fusion outperforms baselines and degrades gracefully.}
\label{fig:hetero}
\end{center}
\vskip -0.2in
\end{figure*}

\section{Experimental Results}
\label{sec:experiments}

\subsection{Simulated experiments}
\label{sec:simulations}

We begin by verifying KL-fusion in a synthetic setting. We consider the problem of fusing Gaussian mixture models with arbitrary means and covariances. Our goal is to estimate true data-generating mixture components by fusing local posterior approximations. 

To quantify estimation quality we compute Hausdorff distances between the polytopes spanned by true and estimated mixture component means, as suggested by \citet{nguyen2013convergence, nguyen2015posterior}. We also evaluate error in the fused model size $G$ relative to the true value. Typical Gaussian mixtures assume a Gaussian--Wishart prior for means and covariances of the components \citep{bishop2006pattern}; in this case, the posterior can be estimated efficiently using mean-field variational inference with Gaussian--Wishart variational distributions \citep{attias2000variational}. To simulate instances of the heterogeneous fusion problem, when generating local dataset, we sample a random subset of the global mixture components and add Gaussian noise to add heterogeneity in model size and parameters. We describe the generating process precisely in the supplemental document.

We consider three baselines: 
\begin{itemize}
    \item ``Oracle'' variational inference (VI) trained on a pooled dataset given the true number of global components (this reference baseline that is infeasible in model fusion);
    \item Dirichlet process (DP) based clustering of the mean components of the local posteriors \citep{ferguson1973bayesian,blei2006variational}; and 
    \item the SPAHM fusion method \cite{yurochkin2019spahm}.
\end{itemize}
The fusion method of \citet{campbell2014approximate} is too inefficient for this problem, since it requires combinatorial search; see \S\ref{sec:fusion-lda} for a comparison to their method.

Our first experiment demonstrates failure of prior methods when the means of the data-generating mixture components are poorly-separated; in this case, we need covariances to disambiguate the components. KL-fusion utilizes full posterior, while SPAHM and DP are limited to only using point estimates of the local posterior means. In Figure \ref{fig:separation}, we vary scale of the variance used to generate means of the true mixture components (higher $x$-axis value implies better separation): KL-fusion is the only fusion method capable of accurate estimation on par with the VI oracle in the low-separation regime. SPAHM performs well only when mixture means are well-separated.

In our second experiment, we fix the separation scale (at 0.5) and study the effect of noise on the generating process for local components. As before, to generate local components we sample from a random subset of the global mixture components and add Gaussian noise to introduce local heterogeneity. We vary this noise in Figure \ref{fig:hetero} and report results on estimating the true number of components as well as the Hausdorff distance between estimated and true global models (in the previous experiment we fixed the noise at 0.5). Intuitively, the problem becomes harder as this noise increases, eventually producing datasets that do not relate to the global structure in a meaningful way. In general, our results with KL-fusion and oracle VI support the intuition that performance degrades gracefully as local models vary more from the underlying global model. Fusion-based inference can even outperform pooled inference: local datasets have less structure and are potentially more amenable to mean-field inference, yielding high-quality approximate posteriors for the subsets of the global model that are easily aggregated using KL-fusion. SPAHM model size estimation error decreases as a function of the noise variance that we add to simulate heterogeneity; this might be caused by the additional separation introduced with this noise.

% \begin{figure*}
% % \vskip 0.2in
% \begin{center}
% \subfigure[Hausdorff distance estimation error]{\includegraphics[width=0.49\textwidth]{figs/lsd_mm-l2.pdf}}
% \subfigure[Global model size estimation error]{\includegraphics[width=0.49\textwidth]{figs/lsd_L-err.pdf}}
% \vskip -0.1in
% \caption{Fusion of Gaussian mixture posteriors under varying heterogeneity in local datasets enforced via noise standard deviation added to global means when simulating local means. KL-fusion outperforms baselines and degrades gracefully.}
% \label{fig:hetero}
% \end{center}
% % \vskip -0.2in
% \end{figure*}

\subsection{Analyzing motion sequences through fused hidden Markov models}

We consider the problem of discovering common structures among related time series. As a motivating application, we study 
data from motion capture sensors on joints of people performing exercise routines, collected from the CMU MoCap database.\footnote{\url{http://mocap.cs.cmu.edu}} Each sequence in this database consists of 64 measurements of human subjects performing various exercises over time. Following~\cite{fox2014joint}, we select 12 informative measurements for capturing gross motor behavior: body torso position, neck angle, two waist angles, and a symmetric pair of right and left angles at each subjects shoulders, wrists, knees, and feet. Each sequence thus is a 12-dimensional time series. We use a curated subset collected by~\citet{fox2014joint} of two subjects each providing three sequences. In addition to having several exercises in common, this subset comes with human-annotated 
labels, facilitating quantitative comparisons between models.

We use mean-field inference with Gaussian--Wishart variational distributions to obtain the approximate posterior of a ``sticky'' HDP-HMM \citep{fox2008hdp}, similar to the analogous experiment by \citet{yurochkin2019spahm}. KL-fusion matches activities inferred from each subject. We use the Rand index \citep{rand1971objective} and Adjusted Mutual Information (AMI) \citep{vinh2010information} to quantify quality of the fused model according the human-annotated labels. Table \ref{table:hmm} compares performance of KL-fusion to SPAHM, demonstrating improved quality of the activity labelings corresponding to the fused models. This experiment gives a practical example where covariances in a Gaussian--Wishart mean-field approximation improve fusion, enabled by KL-fusion's ability to process exponential family distributions.

\begin{table}
\label{table:hmm}
\caption{MoCap labeling quality comparison}
\vskip 0.05in
\begin{center}
\begin{tabular}{lcc}
\toprule
{} &         Rand index & AMI \\
\midrule
KL-fusion &  \textbf{0.286} & \textbf{0.458} \\
SPAHM &  0.254 & 0.445 \\
\bottomrule
\end{tabular}
\end{center}
\vspace{-0.2in}
\end{table}

\subsection{Fusion of topic models}
\label{sec:fusion-lda}
Following \citet{campbell2014approximate}, we run decentralized variational inference on the latent Dirichlet allocation topic model \cite{blei2003latent}. We verify our method against Approximate Merging of Posteriors with Symmetry (AMPS) algorithm from \citet{campbell2014approximate} on the 20 newsgroups dataset, consisting of 18,689 documents with 1,000 held out for testing and a vocabulary of 12,497 words after stemming and stop word removal. 

\begin{figure}[t]
\includegraphics[width=.98\columnwidth]{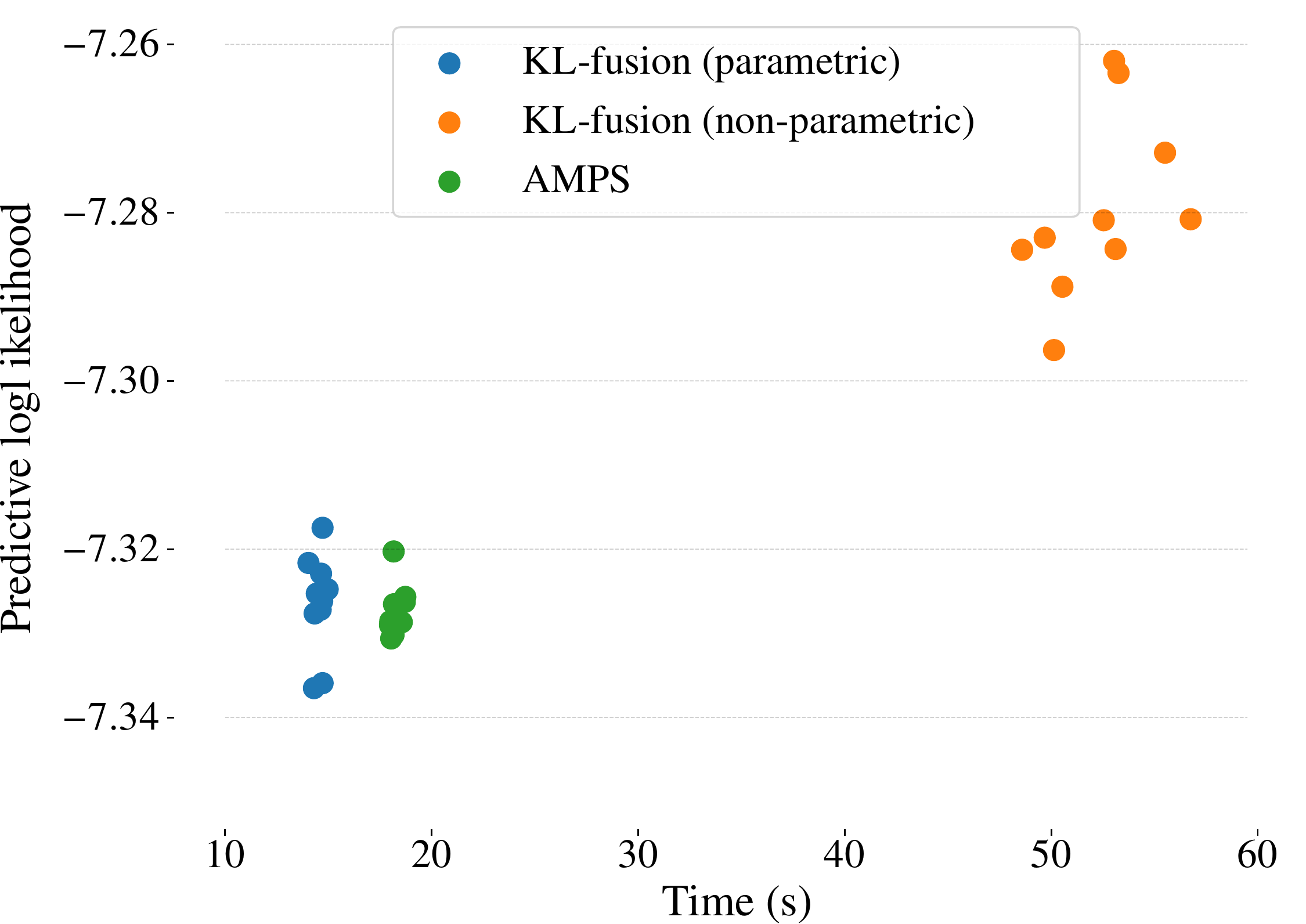}
\vskip -0.1in
\caption{Scatter plot of test $\log$ likelihood and time for the experiment described in \S\ref{sec:fusion-lda}. Higher $\log$ likelihood is better. The parametric model minimizes a very similar objective to AMPS and we observe similar performance between the two. The richer expressiveness of the non-parametric model allows it to perform better, but comes with larger computational requirements.}
\label{fig:lda-fusion}
\vskip -0.2in
\end{figure}

The full description of the model setup is given in the supplemental document. Briefly, the posterior Dirichlet variational parameters learned on the local datasets are fused using KL-fusion and AMPS, and the resulting models are evaluated by computing the predictive $\log$ likelihood of 10\% of the words in each test document given the remaining 90\%. Results are given in Figure \ref{fig:lda-fusion} for $10$ trials; we measure the test $\log$ likelihood and amount of time required to compute the fused model. The parametric model in \eqref{eq:fusion-homogeneous-final} and the objective for AMPS are similar, and thus we expect similar performance from the two methods; our non-parametric version can infer richer models and performs better, but comes at an increased computational cost.

\begin{figure*}[t]
% \vskip 0.2in
\centering
\includegraphics[width= \textwidth]{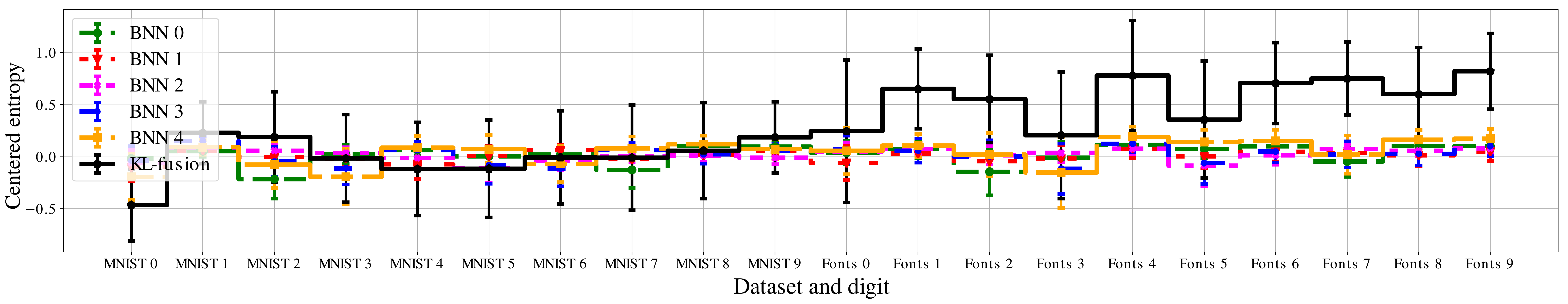}
\vskip -0.1in
\caption{Entropy of local BNNs and KL-fused BNN across 10 digits of MNIST and Fonts centered at the corresponding mean MNIST entropies. Centered entropy above 0 indicates higher uncertainty comparative to that of the MNIST digits for a given BNN. Fused BNN has increased uncertainty over the Fonts digits, while local BNNs do not show such a trend.} 
\label{fig:bnn_step}
\vskip -0.15in
\end{figure*}

\subsection{Fusion of Bayesian neural networks}

In this experiment we demonstrate utility of KL-fusion applied to one-shot federated learning of Bayesian neural networks. Federated learning systems deployed in practice will inevitably face examples outside of the train data distribution leading to mistakes that might be costly for a business or diminish satisfaction of a user with an edge device. Bayesian neural networks are valued for their ability to quantify prediction uncertainty and raise an alarm when facing an out of distribution (OOD) example, but in a federated learning setting data of any individual client might be insufficient to obtain a good quality BNN. 

We measure the effectiveness of our procedure for fusing BNNs locally trained on MNIST digits. For this experiment, we split the MNIST training data into five partitions at random. We simulate a heterogeneous partitioning of the data by sampling the proportion $p_k$ of each class $k$ from a five-dimensional symmetric Dirichlet distribution with a concentration parameter of $0.8$, and allocating a $p_{k,j}$ proportion of the instances of class $k$ to partition $j$. This process results in a non-uniform distribution of classes in each partition. For each dataset we train a single 150-node hidden layer BNN with a horseshoe prior \cite{ghosh2019model, ghosh2018structured}. Horseshoe is a shrinkage prior allowing BNNs to automatically select the appropriate number of hidden units. We use Gaussian variational distribution with diagonal covariance for the weights of the neurons. Details are presented in the supplement.

We use KL-fusion to obtain a global BNN with 281 units, far smaller than a concatenation of all the local BNNs. Table~\ref{table:bnn} illustrates that the fused model significantly improves upon the predictive performance of the local models both in terms of accuracy and held-out test $\log$ likelihoods. We also examine the predictive uncertainties produced by the fused BNN. Figure~\ref{fig:bnn_pics} qualitatively compares the predictive entropy produced by the fused BNN against one of the local BNNs on a set of test images comprising of the standard MNIST test set as well as an out of distribution sample made up of computer generated digits in various fonts \cite{smirnov2020deep}. We observe that while the local network's low entropy samples are heavily influenced by the local training data, the fused model is able to borrow statistical strength from all the local models and exhibits low entropy across all MNIST digits, reserving high entropy predictions for those digits from the OOD set whose fonts look very 
different from the hand drawn MNIST digits.

In Figure \ref{fig:bnn_step} we plot average entropies of the local BNNs and the KL-fused BNN across digits and datasets centering at the corresponding mean entropy on all MNIST digits. The Dirichlet based split of the train data resulted in dataset 0  receiving 2s (78\% of all MNIST images of 2) and 7s (69\%), dataset 4 received 62\% of all 0s and 68\% of 3s (other datasets are also skewed but to a lesser extent). As a result, we observe a lower entropy for 2s and 7s, in both MNIST and Fonts, displayed by BNN 0, and similar for BNN 4 on 0s and 3s. This observation suggests (as well as Figure \ref{fig:bnn_pics}) that these BNNs have learned to distinguish their dominant digits from the rest of the digits, rather than the desired BNN classifying all 10 digits and being uncertain on OOD examples. Our KL fusion method is able to produce a BNN with the desired properties from these biased local BNNs as entropy increases significantly on the OOD Fonts digits.

\begin{table}
\vskip -0.1in
\caption{Comparison of local and fused BNNs}
% \vspace{-10pt}
\label{table:bnn}
\vskip 0.05in
\begin{center}
\begin{tabular}{lccc}
\toprule
{} &         Accuracy, \% &  Test ll & Entropy \\
\midrule
KL-fusion &  \textbf{95.8} & \textbf{-0.32} & \textbf{0.79} \\
BNN 0 &  90.1 & -1.42 & 2.12 \\
BNN 1 &  91.6 & -1.44 & 2.13 \\
BNN 2 &  82.7 & -1.56 & 2.17 \\ 
BNN 3 &  87.8 & -1.37 & 2.08 \\
BNN 4 &  91.9 & -1.18 & 2.00 \\
\bottomrule
\end{tabular}
\end{center}
\vskip -0.25in
\end{table}
\section{Conclusion}

Federated learning techniques vary in complexity and communication overhead.  On one extreme, some approaches hand information back and forth between different entities as they reach a consensus on the global model.  On the other extreme, model fusion extracts a global model in a single shot:  Local models are combined into the global model by solving a single optimization problem, and then the learning procedure is complete.  Our technique and experiments show that model fusion can be effective despite its simplicity:  In a single step, we extract a global model capturing relevant information from multiple local models.

The design of an effective fusion algorithm combines several key ideas.  Working specifically with mean field approximations and exponential family distributions leads to a feasible algorithm while staying applicable to a wide array of practical scenarios, as illustrated by the examples in \S\ref{sec:experiments}. This setup also allows our method to use information about the full local distributions, rather than point estimates as in previous work \cite{yurochkin2019spahm}.  Moreover, mixed norm regularization dynamically adjusts the dimensionality of our fused model.  More broadly, model fusion in the Bayesian setting accompanies the fused model with uncertainty estimates, valuable for detecting out-of-distribution samples that are not captured by any of the individual local models as in Figure \ref{fig:bnn_pics}.

The success of KL-fusion suggests several avenues for future work.  We likely can extend our algorithm to Bregman divergences other than KL using a similar formulation and algorithm; farther afield, optimal transport distances could improve the quality of our inference procedure but would likely require adjustment to the simplifications outlined in \S\ref{sec:homogeneous-fusion}.  We also could extend our method to handle \emph{iterative} refinement, communicating the global model back to the local models as a means of improving the analysis of each component dataset.

KL-fusion seeks to find a global model to best approximate all of the local posteriors. In the distributed posterior inference literature the goal is often to approximate the posterior distribution of a pooled dataset assuming homogeneous data partition, e.g. \citet{srivastava2015wasp}. Understating the connection between fused model and pooled data posterior is an interesting theoretical problem demanding new proof techniques to account for permutation invariance and data heterogeneity considered in our KL-fusion algorithm. 

\paragraph*{Acknowledgments.} Justin Solomon and the MIT Geometric Data Processing group acknowledge the generous support of Army Research Office grants W911NF1710068 and W911NF2010168, of Air Force Office of Scientific Research award FA9550-19-1-031, of National Science Foundation grant IIS-1838071, from the MIT--IBM Watson AI Laboratory, from the Toyota--CSAIL Joint Research Center, from a gift from Adobe Systems, and from the Skoltech--MIT Next Generation Program.

\bibliography{ref}
\bibliographystyle{icml2020}

\clearpage
\appendix

\section{Simulated experiments details}
In \S\ref{sec:simulations} we studied fusion of mixture model posteriors learned from heterogeneous datasets. Below we describe the data generating process used in these experiments.

First, we generate true global means $\mu_g \sim \mathcal{N}(0,I_D\sigma_0^2) \in \mathbb{R}^D$ for $g=1,\ldots,G$. We set true number of global components $G=5$, data dimension $D=10$, and entries of the diagonal covariance $\sigma_0^2 = sG$. Parameter $s$ is the separation scale controlling the degree of separation between the true global means and corresponds to the $x$-axis in Figure \ref{fig:separation}. To generate covariances $\{\Sigma_g\}_{g=1}^G$ of the global mixture components we used a slightly modified Scikit-learn \citep{pedregosa2011scikit} function for random positive definite matrices (see code for details).\footnote{Code link: \url{https://github.com/IBM/KL-fusion}}
% $A_{ij} \sim \mathcal{N}(0,1)$ for $i,j=1,\ldots,D$; $U, \Lambda, V = \text{SVD}(A^T A)$, $\Lambda_{ii} = \text{min}(\Lambda_{ii}, 1)$ for $i=1,\ldots,D$; $\Sigma_g = U (\Lambda + I_D g) V$.

\begin{figure}
\includegraphics[width=.98\columnwidth]{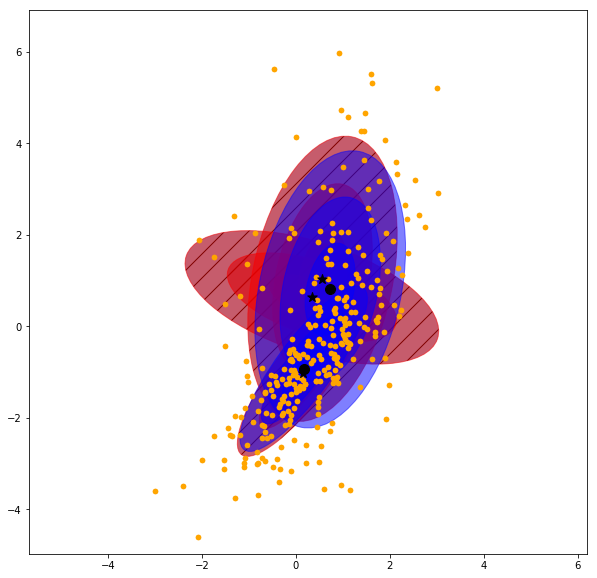}
\caption{Illustration of the data simulation process in two dimensions: contours of the Gaussian densities corresponding to the global mixture components are in red with black stars as means. Components of the local mixture model are in blue with black circles marking means: our data generative process selected a subset of two global mixture components and perturbed means with some Gaussian noise to enforce heterogeneity. Corresponding local dataset is shown in orange.}
\label{fig:data_gen}
\vskip -0.2in
\end{figure}

To generate $J=50$ heterogeneous datasets, we first assign a random probability to each global component. For each dataset $j$ we then select a random subset of global means and covariances by drawing Bernoulli random variables with the corresponding probabilities. This corresponds to the Beta-Bernoulli process \citep{thibaux2007hierarchical}. Next, to each selected mean we add Gaussian noise with standard deviation $\sigma$ to enforce heterogeneity in the parameters. This $\sigma$ is the $x$-axis in Figure \ref{fig:hetero}. We also slightly perturb covariances using Wishart distribution (see code for details). Finally, each dataset is generated from a Gaussian mixture with the corresponding means and covariances and mixture component probabilities drawn from a symmetric Dirichlet distribution. To illustrate the generative process for one dataset, in Figure \ref{fig:data_gen} we give an example in $D=2$ dimensions, with $G=3$ global components, separation scale $s=0.1$ and heterogeneity noise $\sigma=0.1$. Our generative process resulted in a local mixture model with two components slightly perturbed from the corresponding global components.

\begin{figure*}[t]
% \vskip 0.2in
\begin{center}
\subfigure[Clustering iteration of KL-fusion: after several iterations our algorithm learned meaningful clustering of the distributions corresponding to the posterior approximations of the local mixture components. Distributions in the same cluster are shown with the same color and mean marker. Our algorithm correctly identified that there should be 3 clusters.]{\includegraphics[width=0.48\textwidth]{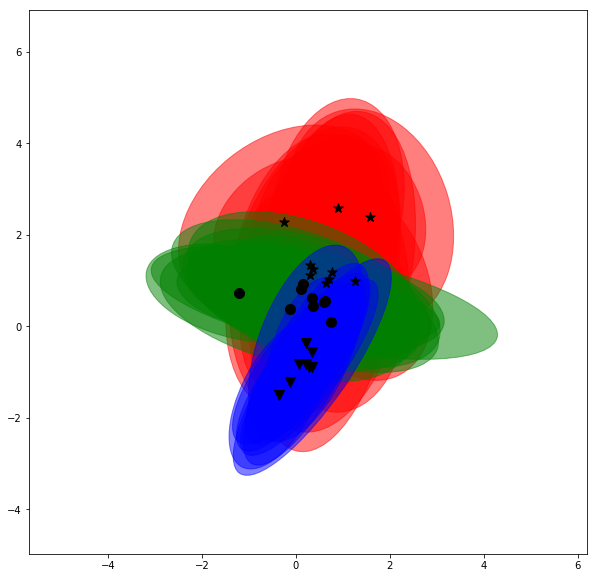}}
\hfill\subfigure[Fused posterior learned with KL-fusion: in blue we show estimates of the fused mixture model means and covariances obtained using our algorithm. This is the result of taking KL barycenter corresponding to the clustering on the left. In red we show true global means and covariances. KL-fusion produces an accurate estimate and estimates the size of the global model correctly.]{\includegraphics[width=0.48\textwidth]{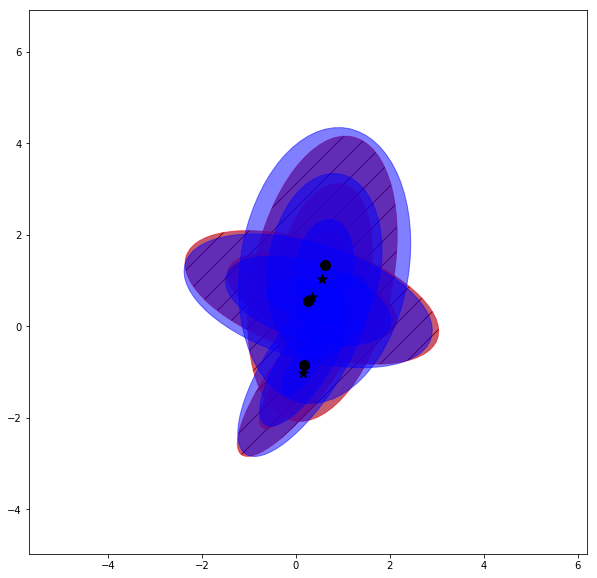}}
\vskip -0.1in
\caption{Visualization of the KL-fusion algorithm in 2 dimensions}
\label{fig:toy_fusion}
\end{center}
% \vskip -0.2in
\end{figure*}

To obtain local posteriors for KL-fusion, on each of the datasets we ran variational inference with Gaussian-Wishart variational distributions.\footnote{\emph{sklearn.mixture.BayesianGaussianMixture}} Recall that our KL-fusion algorithm alternates between clustering of the components of the local posterior distributions and finding corresponding barycenters. In Figure \ref{fig:toy_fusion}(a) we show the result of the clustering step on one of the iterations of KL-fusion and in Figure \ref{fig:toy_fusion}(b) we present estimates of the fused mixture model means and covariances obtained from the corresponding barycenter.

\section{Topic modeling details}
The problem setup for the topic modeling experiment was as follows: The 20news training dataset was split into 5 separate datasets based on topics of the news articles. For each local dataset, we fit a topic model with 10 topics. In this context, each component is a topic, the parameters for each component are the posterior Dirichlet variational parameters (one for each word), and the goal of fusion is to infer these posterior Dirichlet parameters for the global model. We then fused these posterior distributions using either AMPS \cite{campbell2014approximate}, the parametric version of our algorithm, or the non-parametric version of our algorithm.

For each document in the test set, we measure predictive log likelihood of the $10\%$ of the words based on the remaining $90\%$. Predictive log likelihood is defined as in \citet{wang2011online}, and approximated by
\begin{equation}
    p(\textbf{w}_{j2} \,|\, \textbf{w}_{j1}, \mathcal{D}_{\mathrm{train}}) = \prod_{w\in \textbf{w}_{j2}}\sum_k \bar{\pi}_{jk}\bar{\phi}_{kw}
\end{equation}
where $\bar{\pi}_{jk}$ is the proportion of topic $k$ in document $j$, and $\bar{\phi}_{kw}$ is the posterior Dirichlet parameter of word $w$ in topic $k$, $\textbf{w}_{j2}$ are the held-out words. The $\log$ of this quantitity is summed over all documents in the test set. 

\section{HDP-HMM details}
Our HMM models use multivariate Normal-Wishart observation models and Hierarchical Dirichlet process allocation models. The state specific transition probabilities $\pi_k$ are drawn according to the following generative process. First, we draw $\beta \sim \text{GEM}(\gamma)$ from the stick breaking distribution. That is,
\begin{equation}
    \beta_j = \nu_j\prod_{l=1}^{j-1} (1 - \nu_l); \quad \nu_j \mid \gamma \sim \text{Beta}(1, \gamma); \quad j = 1, 2, \ldots,
\end{equation}
We then draw $\pi_k$ from a Dirichlet process with a discrete base measure shared across states, 
\begin{equation}
\pi_k \mid \eta, \kappa, \beta \sim \text{DP}(\eta + \kappa, \frac{\eta\beta + \kappa\delta_k}{\eta + \kappa}); \quad k = 1, 2, \ldots,
\end{equation}
where $\eta$ is a concentration parameter and $\kappa$ is a ``stickyness'' parameter~\cite{fox2008hdp} that encourages state persistence. 
The latent states for a particular sequence then evolve as $z_t$, evolve as $z_{t+1} \mid z_t, \{\pi_k\}_{k=1}^{\infty} \sim \pi_{z_t}$. Finally, observations at time step $t$, $y_t \in \mathbb{R}^D$ are drawn from a Normal Wishart distribution,
\begin{equation}
\begin{split}
    \mu_{k} \mid \mu_0, \lambda, \Lambda_k &\sim \mathcal{N}(\mu_0, (\lambda\Lambda_k)^{-1}) \\
    \Lambda_{k} \mid S, n_0 &\sim \text{Wishart}(n_0, S) \\
    y_t \mid z_t=k &\sim \mathcal{N}(y_t \mid \mu_{k}, \Lambda_{k}^{-1})
\end{split}
\end{equation}    
For all our experiments, we set $\kappa$ to 10.0, $\gamma = 5.$ and $\eta=0.5$. For the observation model, we set $n_0 = 1$ and $S$ to an identity matrix $I$, encoding our belief that $\mathbb{E}[\Lambda_k^{-1}] = I$.    

\subsection {MoCAP data details}
We consider the problem of discovering common structure in collections of related time series. Although such problems arise in a wide variety of domains, here we restrict our attention to data captured from motion capture sensors on joints of people performing exercise routines. We collected this data from the CMU MoCap database (\url{http://mocap.cs.cmu.edu}). Each motion capture sequence in this database consists of 64 measurements of human subjects performing various exercises. Following~\citet{fox2014joint}, we select 12 measurements deemed most informative for capturing gross motor behaviors: body torso position, neck angle, two waist angles, and a symmetric pair of right and left angles at each subjects shoulders, wrists, knees, and feet. Each MoCAP sequence thus provides a 12-dimensional time series. We use a curated subset~\cite{fox2014joint} of the data from two different subjects each providing three sequences. In addition to having several exercise types in common this subset comes with human annotated labels allowing for easy quantitative comparisons across different models.

\section{Bayesian Neural Network details}
We use Bayesian neural networks with regularized horseshoe priors~\cite{ghosh2019model, ghosh2018structured} as our local BNN models. In more detail, let a network with $L-1$ hidden layers be parameterized by a set of weight matrices $\W{} = \{W_l\}_{1}^{L}$, where each weight matrix $W_l$ is of size $\real{(K_{l-1} + 1)\times K_{l}}$, and $K_l$
is the number of units (excluding the bias) in layer $l$.  Let the node weight vector $\wnode \in \real{(K_{l-1}+1)}$ denote the set of weights incident to unit $k$ of hidden layer $l$. Following \citet{ghosh2019model, ghosh2018structured}, we place regularized horseshoe priors over $\wnode$:
\begin{equation}
	\wnode \mid \taunode, \taulayer, c \sim \mathcal{N}(0, (\tildetaunode^2\taulayer^2)I),\quad  \tildetaunode^2 = \frac{c^2\taunode^2}{c^2 + \taunode^2\taulayer^2},
	\label{eq:reg-HS}
\end{equation}
with $\taunode \sim \halfcauchy(0, \bnode), \quad \taulayer \sim \halfcauchy(0, \blayer)$, and $c \sim \invgamma(c_a, c_b)$. Here, $I$ is an identity matrix and $a \sim \halfcauchy(0, b)$ is the half-Cauchy distribution with density $p(a|b) = 2/\pi b(1 + (a^2/b^2))$ for $a>0$; $\taunode$ is a unit specific scale parameter,
while the scale parameter $\taulayer$ is shared across layer $l$. 
The regularized horseshoe distribution exhibits a spike at zero that provides strong shrinkage towards zero and encourages sparsity by turning off nodes in a layer that are not necessary for explaining the data. This allows the local BNNs to automatically select the appropriate number of nodes in each layer.  

We resort to variational inference on an equivalent parameterization, $\wnode = \tildetaunode\taulayer\betanode, \quad \betanode \sim \normal(0, I)$.
After learning the variational posteriors of this equivalent model, we arrive at $q(\wnode)$ by first approximating the factorized variational approximations $q(c), q(\taunode), q(\taulayer)$ with their expected values $\mu_c, \mu_{\taunode}, \mu_{\taulayer}$, and defining
$$
\mu_{\tildetaunode}^2 = \frac{\mu_c^2\mu_{\taunode}^2}{\mu_c^2 + \mu_{\taunode}^2\mu_{\taulayer}^2}.
$$
Given the expected values and the Gaussian variational distribution $q(\betanode) = \normal(\mu_{\betanode}, \Psi_{\betanode})$,  $\wnode = \tildetaunode\taulayer\betanode$ follows a Gaussian distribution $\normal(\mu_{\wnode}, \Sigma_{\wnode})$, with
$$
\mu_{\wnode} = \mu_{\tildetaunode}\mu_{\taulayer}\mu_{\betanode}; \quad \Sigma_{\wnode} = \mu_{\tildetaunode}^2\mu_{\taulayer}^2\Psi_{\betanode}.
$$
We thus recover the variational approximation on $\wnode$, $q(\wnode) = \normal(\mu_{\wnode}, \Sigma_{\wnode})$. These variational distributions from different local BNNs are then used for fusion.

\section{Initialization}
To initialize KL-fusion algorithm we used a variation of $k$-means++ initialization as discussed in the conclusion of \citet{arthur2006k}. This is a popular initialization scheme used in Scikit-learn \citep{pedregosa2011scikit} for $k$-means clustering. In our KL-fusion initialization, we replaced squared Euclidean distance with $\mathrm{KL}$ divergence.

\end{document}